\documentclass{article}
\usepackage[utf8]{inputenc}
\usepackage[font=small,labelfont=bf,margin=0.3cm]{caption}
\usepackage{textcomp}
\usepackage{gensymb}
\usepackage{titling}
\usepackage{lipsum}
\usepackage{verbatim}
\usepackage{amsmath}
\usepackage{amsfonts}
\usepackage{float}
\usepackage{tikz}
\usepackage{graphicx}
\usepackage{subcaption}

\usetikzlibrary{shapes.geometric, arrows}

\tikzstyle{startstop} = [rectangle, rounded corners, minimum width=3cm, minimum height=1cm,text centered, draw=black, fill=green!20]
\tikzstyle{io} = [trapezium, trapezium left angle=70, trapezium right angle=110, minimum width=3cm, minimum height=1cm, text centered, draw=black, fill=blue!30]
\tikzstyle{process} = [rectangle, minimum width=3cm, minimum height=1cm, text centered, draw=black, fill=orange!30]
\tikzstyle{decision} = [diamond, minimum width=3cm, minimum height=1cm, text centered, draw=black, fill=green!30]
\tikzstyle{arrow} = [thick,->,>=stealth]
\usepackage[linesnumbered,ruled]{algorithm2e}
\usepackage[margin=1in]{geometry} 
\newcommand{\cmmnt}[1]{}

\pretitle{\begin{center}\huge\bfseries}
\posttitle{\par\end{center}\vskip 0.5em}
\preauthor{\begin{center}\Large}
\postauthor{\end{center}}
\predate{\par\large\centering}
\postdate{\par}

\title{Change-point Detection Methods for Body-Worn Video} 
\author{Stephanie Allen\footnote[2]{Department of Mathematics, State University of New York at Geneseo}, David Madras\footnote[3]{Department of Computer Science, University of Toronto}, Ye Ye\footnote[4]{Department of Mathematics, University of California, Los Angeles}, Greg Zanotti\footnote[5]{Department of Mathematics/Computer Science, DePaul University}\footnote[1]{Names ordered alphabetically to signify equal contribution to the work by all authors.}}
\date{\today}
\begin{document}

\maketitle
\thispagestyle{empty}

\begin{abstract}
Body-worn video (BWV) cameras are increasingly utilized by police departments to provide a record of police-public interactions.  However, large-scale BWV deployment produces terabytes of data per week, necessitating the development of effective computational methods to identify salient changes in video.  In work carried out at the 2016 RIPS program at IPAM, UCLA, we present a novel two-stage framework for video change-point detection.  First, we employ state-of-the-art machine learning methods including convolutional neural networks and support vector machines for scene classification.  We then develop and compare change-point detection algorithms utilizing mean squared-error minimization, forecasting methods, hidden Markov models, and maximum likelihood estimation to identify noteworthy changes.  We test our framework on detection of vehicle exits and entrances in a BWV data set provided by the Los Angeles Police Department and achieve over 90\% recall and nearly 70\% precision --- demonstrating robustness to rapid scene changes, extreme luminance differences, and frequent camera occlusions. 
\end{abstract}
\section{Introduction}\label{introduction}
Body-worn video (BWV) cameras are becoming increasingly popular tools for police departments \cite{miller2014implementing}. They are used to provide a record of police-public interactions, and have been shown to increase accountability among officers \cite{katz2014evaluating}. Furthermore, BWV has recently become a topic of widespread interest among the general public, especially given the recent controversies regarding police-public relations and policy. To produce this video, police officers wear specially designed cameras on their chests to record their interactions with the public. 
However, large-scale BWV deployment produces terabytes of data per week, far too much for complete review by humans. This necessitates the development of effective computational methods to identify salient changes in video between various states --- such as in or out of a building, interacting or not interacting with the public, and in and out of a car.  

In early architectures in the literature, changes in videos are detected using a variety of statistical and image processing techniques based on computing differences in image feature representations \cite{vidcammotion}. 
Other methods extend these basic spatiotemporal models in interesting ways to produce video-specific change-point detection algorithms. For example, in \cite{pliss}, the authors introduce a Bayesian method to segment videos containing specific scenes into clusters in an online, unsupervised way accompanied by confidence probabilities. The method is applied to robotics.  In \cite{robustonline2006}, the authors extend a statistical change-point detection algorithm to video in order to track 3D objects. In recent deep learning literature, the authors in  \cite{Karpathy_2014_CVPR} propose a convolutional network with a sliding frame window input capable of creating spatiotemporal features to classify videos.

The change-point detection literature informs part of our approach as well. Classic statistical methods range from simple sum- and mean-based thresholding algorithms for single change-point detection in offline data \cite{chen2001change}, to nonparametric tests for changes in distributions \cite{kstest-twosample}. 
Other statistical methods use Bayesian priors to incorporate time-dependent information into the probability of a change-point occurring \cite{adams2007bayesian}.

In this paper, we present a novel two-stage framework (summarized in Figure 1) for video change-point detection which draws on methods from machine learning, computer vision, and change-point detection.  We begin with a video data set with ground truths --- the time at which changes between the two predefined states occur.
These states are mutually exclusive and collectively exhaustive, and we refer to them as positive and negative states. Then, we selectively extract frames from each video to create a time series of frames. In the first stage, we utilize feature extraction and image representation methods to generate a compact representation of each frame and then label each representation via classifiers --- support vector machine (SVM) and convolutional neural network (CNN) --- and we ultimately construct a time series of scores. These scores measure the confidence of a classifier about whether a video frame corresponds to the positive state. In addition, by setting a threshold, we are able to convert these scores into binary labels (0,1) corresponding to positive and negative states. Finally, change-point detection algorithms analyze the scores or labels to identify salient changes between the two states of interest, thereby locating the times at which change-points occur. This modular format enables generalization to a variety of change-point classes.      

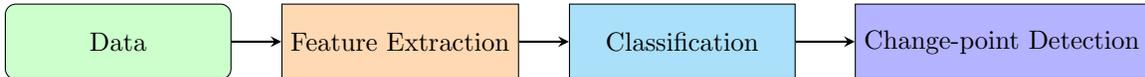
\begin{figure}[H]
	\begin{center}
    \begin{tikzpicture}[node distance=1.75cm]
		\node (start) [startstop] {Data};
        \node (pro1) [process, right of=start, xshift=2cm] {Feature Extraction};
		\node (pro2a) [process, fill=cyan!30, right of=pro1, xshift=2cm] {Classification};
        \node (pro3) [process, fill=blue!30, right of=pro2a, xshift=2.5cm] {Change-point Detection};
        \draw [arrow] (start) -- (pro1);
        \draw [arrow] (pro1) -- (pro2a);
		\draw [arrow] (pro2a) -- (pro3);
    \end{tikzpicture}
    \end{center}
\caption{Our framework's workflow for video change-point detection.}
\end{figure}

The paper is organized as follows. Section~\ref{video_preprocessing} presents construction of video representation and classification approaches, turning to the computer vision literature using feature detection methods, SVMs, 
and CNNs 
Change-point detection methods are presented in Section~\ref{change-point methods}, utilizing mean squared-error minimization, forecasting methods, hidden Markov models, and maximum likelihood estimation. 
Finally, we perform an experiment on a body-worn video data set provided by the Los Angeles Police Department (LAPD). We parameterized our framework to detect changes from in-car scenes to out-of-car scenes, and we achieve promising results, which are presented in Sections~\ref{classification results} and \ref{change-point results}.

\section{Video Preprocessing and Frame Classification}\label{video_preprocessing}
A video can be regarded as a sequence of frames. In a preprocessing step, we sample frames from videos and save them as JPEG images. The goal is to classify these frames as either one of the two states --- the states between which we wish to identify change-points. 
We frame this problem as one of scene classification. Scene classification has been extensively studied by the computer vision community; consequently we use methods from computer vision to classify scenes. 
Current state of the art approaches use either keypoint detection and the Bag-of-Visual-Words (BoVW) technique with a classifier (such as an SVM) capable of comparing the histograms it produces \cite{yang2007evaluating}, or a convolutional neural network (CNN) on raw pixel values \cite{alexnet}. To extend the SVM, we propose a novel technique for soft histogramming, which improves classification accuracy. We also modify the architecture of a pre-trained CNN to create a CNN capable of two-state video frame classification. The details are described in the following sections. 

\subsection{Keypoint Detection and Support Vector Machine}\label{keypoint_and_svm}
Intuitively, keypoints are distinctive image features. After a keypoint is located by a keypoint detector, image features in the keypoint's neighborhood can be described by a keypoint descriptor. 
We use scale-invariant feature transform (SIFT) \cite{lowe2004distinctive} for keypoint detection and description, because SIFT features are shown to be invariant to image scale and rotation, and it is partially invariant to changes in illumination. The major steps of constructing SIFT can be summarized as:

\begin{itemize}
\item Keypoint detection and localization 
  \begin{itemize}
  \item Apply Gaussian filters with different standard deviations to the input frame
  \item In the differences of Gaussians, search for local extrema in scale and space. These extrema are potential keypoints.
  \end{itemize}
 \item Orientation assignment
  \begin{itemize}
  \item Assign one or more orientations to each keypoint based on the directions of pixel gradients in the keypoint's neighborhood
  \end{itemize}
\item Keypoint description
  \begin{itemize}
  \item Compute gradients of pixels relative to the keypoint orientation in the 16-by-16 neighborhood around each keypoint
  \item Divide this 16-by-16 patch into blocks of 4-by-4 in size and create an 8-bin orientation histogram for each block
  \item Concatenate histograms to get a 128 dimensional descriptor for each keypoint
  \end{itemize}
\end{itemize}

Using this approach, each frame can be represented as a SIFT matrix, with each row being a 128-dimenional SIFT descriptor. However, since the number of SIFT descriptors extracted varies among frames and an SVM requires inputs to have the same dimension, we use BoVW as an additional step to construct image representations.
\subsubsection*{Bag-of-Visual-Words and Vector Quantization}
After extracting SIFT features from all video frames of interest, we took 20\% of frames of the two states in the training set and applied \textit{k}-means clustering \textit{separately} on their feature vectors, with each state having $K$ clusters. 
After centroids of clusters are computed, we assign each feature vector from frames in testing set and the remaining part of the training set to its closet centroids based on Euclidean distance. This general technique is called BoVW, and the number of clusters $K$ is often referred to as the size of vocabulary. BoVW is an example of vector quantization (VQ) in computer vision. After VQ, a feature vector is represented by the indices of clusters to which it is assigned \cite{vq}. In general, there are two distinct forms of VQ: \textit{hard} VQ and \textit{soft} VQ. In hard VQ, a feature vector is assigned to exactly one cluster, which corresponds to the closest centroid; whereas in soft VQ, a feature vector can be assigned to more than one clusters \cite{viitaniemi2009spatial}. In our work, a feature vector's membership at each cluster depends on the feature vector's distance to the corresponding centroid. We propose the following technique to perform soft VQ.

Let $\{c_{j}\}^C_{j=1}$ be a set of centroids computed in the clustering stage, where $C=2K$ is the total number of centroids, and let $\{f_{i}\}^F_{i=1}$ denote the set of all SIFT feature vectors extracted from a frame. The goal is to construct $H\in\mathbb{R}^{C}$, where $H_{j}$ measures the effective number of feature vectors assigned to cluster $j$ for $1\le j \le C$. For each feature vector $f_{i}$, we compute its Euclidean distance $D_{ij}$ to each of the centroids $c_{j}$. Then, the \textit{relative} distance between centroid $c_{j}$ and feature vector $D_{ij}$ can be defined as $$R_{ij}=\frac{D_{ij}-\min\limits_{1\le p\le C}(D_{ip})}{\max\limits_{1\le p\le C}(D_{ip})-\min\limits_{1\le p\le C}(D_{ip})},$$ where the centroid closest to the $f_{i}$ has relative distance $0$ whereas the farthest centroid gets relative distance $1$. To control the contribution of $f_{i}$ to clusters whose corresponding centroids are not the closest to $f_{i}$, a parameter $E$ is introduced. We then define the exponentially decayed relative distance $R'_{ij}$ as $R'_{ij}=\exp\left(-ER_{ij}\right)$ so that we essentially recover hard VQ as $E$ approaches positive infinity. 
The contribution of $f_{i}$ to $H$ is then normalized to 1, and so every feature vector has the same weight. This procedure is summarized in Algorithm~\ref{soft_vq} below.

Note that the idea of VQ is closely related to histogramming: assigning a vector to clusters essentially achieves the same effect as incrementing the counts at the corresponding histogram bins, except that in the later case bin counts are discrete. For notational convenience, we refer to soft VQ and soft histogramming interchangeably, and $H$ is called ``BoVW histogram'' in subsequent sections.
\begin{algorithm}
\caption{Soft VQ}\label{soft_vq}

\SetKwInOut{Input}{Inputs}
\SetKwInOut{Output}{Output}
\Input{set of centroids $\{c_{j}\}^C_{j=1}$,\\ set of feature vectors $\{f_{i}\}^F_{i=1}$ from an frame,\\ a positive constant $E$}
\Output{$H\in\mathbb{R}^{C}$}
For $i=1 : F$\\
\quad For $j=1 : C$\\
\qquad\quad $D_{ij}\gets ||f_{i}-c_{j}||_{2}$\\

\quad For $j=1:C$\\
\qquad\quad $R_{ij}\gets\frac{D_{ij}-\min\limits_{1\le p\le C}(D_{ip})}{\max\limits_{1\le p\le C}(D_{ip})-\min\limits_{1\le p\le C}(D_{ip})}$\\
\qquad\quad $R'_{ij}\gets\exp\left(-ER_{ij}\right)$\\
\quad For $j=1:C$\\
\qquad\quad $H_{j}\gets H_{j} + \frac{R'_{ij}}{\sum\limits^C_{p=1}R'_{ip}}$\\
\end{algorithm}

Note that conventional BoVW and VQ do not consider spatial information of keypoints. In other words, the locations of objects within images are not taken into consideration. Previous literature \cite{yang2007evaluating} suggests that for a small size of visual vocabulary, including spatial information can improve classifiers' performances significantly, while for a large size of visual vocabulary, the improvements are not substantial.   
\subsubsection*{Support Vector Machine with Pyramid Match Kernel} 

To evaluate the effect of including spatial information of keypoints, we experiment with pyramid match kernel \cite{lazebnik2006beyond}, which partitions an input image into increasingly fine spatial bins. At level $l$, $2^l$ cells are placed on each side of an image, so there are $4^l$ spatial bins with equal size. No partition occurs at level $0$. By setting parameter $L$, which is the maximum number of levels, we are able to control how much detailed spatial information are included. For example, if $L$ is set to 0, no spatial information of keypoints will be considered. 

After VQ, we can represent a feature vector by a histogram of cluster membership, where each histogram bin measures the similarity between the feature vector and the corresponding centroid. For each spatial bin, we create an aggregated histogram by summing up histograms corresponding to feature vectors falling in this bin. These aggregated histograms are then weighted according to the level at which the spatial bin is located. Because matches of features at finer spatial resolutions are expected to yield more information about the similarity between two images, histograms at finer grids are weighted more heavily. We follow the practice in \cite{lazebnik2006beyond} and give weights  1/4, 1/4, and 1/2 to levels 0, 1, and 2 respectively. In the final step, we concatenate these weighted aggregated histograms from all spatial bins, and an input image is represented by a vector of a fixed length. This vector is later input into SVM.

A two-class SVM works by finding the optimal hyperplane that gives the maximum separation between training examples from the two classes. This goal can be achieved by solving an optimization problem, which maximizes the two-class separation while penalizing training examples lying on wrong sides of margins. We apply a kernel SVM, which first maps training examples to a higher dimensional feature space before optimizing for the maximum separation. A kernel function takes two training examples and measures their similarity. In our project, the kernel function corresponds to the pyramid match kernel \cite{lazebnik2006beyond}, which sums up the intersections between two histogram created using the method described above.

\subsection{Convolutional Neural Network} \label{SS:CNN}
In the second classification approach, we use deep neural networks. Deep neural networks are machine learning algorithms that jointly learn a feature representation and discriminative classifier over a data set \cite{Goodfellow-et-al-2016-Book}. Nonlinear computational nodes called \textit{neurons} are stacked on top of one another in layers to form complex, richly informative sets of features that have highly discriminative characteristics. Neural networks are trained by changing weights, thresholds, and other parameters, generally through the use of an iterative optimization algorithm like stochastic gradient descent \cite{Goodfellow-et-al-2016-Book,rumelhart1985learning}. An overview of the historical development of neural networks and deep learning can be found in \cite{schmidhuber2015deep}.

Convolutional neural networks (sometimes referred to as ``ConvNets'', ``convolutional networks'', or ``CNNs'') have their origins in the study of the visual cortex in primates \cite{receptive}. ConvNets were popularized in \cite{lecun1989}, although earlier forerunners such as \cite{neocognitron} contributed to their development. ConvNets extract information from input data using overlapping convolutions. Each convolution operation consists of "sliding" a feature detector over input data, which generates an output of similar dimensionality. Each feature detector looks for one specific feature, and is made up of a number of trainable weights. Features are dependent on data; for example, image features may include edges, color blobs, or simple shapes. Multiple convolutions are performed in a single \textit{convolutional layer}, and the output of a convolutional layer is  transformed by nonlinear activation functions. Convolutional layers are stacked and interspersed with \textit{pooling layers}, which subsample their input (e.g.\ by taking averages over various input sections) and produce an output with lower dimension.


A convolutional layer is made up of several different feature maps, which take the form of tensors, in the sense that they are (small) multi-dimensional arrays of numbers. The feature maps require this definition because the two-dimensional input image requires the first convolutional layer to contain two-dimensional feature maps. Because there are multiple feature maps in the first layer, the output matrices are concatenated to form output tensors---three-dimensional arrays that are convolved with the feature maps in the next convolutional layers. This structure means that convolutional networks pass tensors in between their intermediate layers. 

At the last pooling or convolutional layer, the produced activation tensor is generally flattened into a single vector, and connected to a fully connected layer. This fully connected layer is followed by one or two more fully connected layers, and an output layer. In the case of binary classification, we measure the error of the output layer activation (or ``score'') using the \textit{hinge loss} function. Hinge loss is defined as $\ell(y, \hat{y}) = \max (0, 1-y \cdot \hat{y})$, where $y$ is the true label and $\hat{y}$ is the predicted score. The combination of convolutional layers, which create high-quality, deep feature representations of images, and fully connected neural networks, which are excellent classifiers, make convolutional neural networks very effective at most computer vision tasks.

\subsection{Pre-trained Networks}
Unfortunately, deep neural networks (especially convolutional networks) have one negative feature: they take large amounts of time and computational power to train. One way to bypass this problem is to re-use a popular, well-known network configuration for which trained weights already exist. These pre-trained neural networks are created by neural network researchers, and consist of a known architecture (e.g. a number of convolutional layers, specific sizes of feature maps, etc.), and a file containing the numbers for each weight in the network. 

Pre-trained networks have been released for ILSVRC, a competition in which participants aim to classify images into one of 1,000 classes. It has been found that the weights and structure of these networks provide excellent starting points for classifying images into a different set of classes. For example, in \cite{Oquab_2014_CVPR}, it is reported that simply removing the output layer of a popular pre-trained ConvNet and replacing it with a new output layer trained to detect a different set of classes provides state of the art accuracy on several computer vision problems. 

\subsubsection*{VGG-16 Architecture}
The pre-trained VGG-16 network of \cite{vggnet} was initially conceived and publicly released in 2014. It was a top-performing model in the 2014 ILSVRC, and has been used widely in the literature to achieve excellent results on image classification problems. The original training process of the VGG-16 network can be found in \cite{vggnet}. The main reason we use the VGG-16 convolutional network is for its very deep architecture, which is what allowed it to perform so well in the 2014 ILSVRC competition.

\subsubsection*{Adapting VGG-16}
Although VGG-16 ends with a 1,000-dimensional output layer, this layer can be removed and replaced with a layer made for a binary classification task such as detecting whether a scene is classified into one state or another. To facilitate this, the weights were downloaded from the authors' website, and the network was implemented using machine learning software libraries. Because the weights of the last fully connected layer are often tuned to the task of the next (output) layer \cite{Oquab_2014_CVPR}, we removed this layer as well as the output layer. We replaced the two layers with a single output layer that uses the hinge loss function. This new layer, once the weights are trained, produces a univariate scalar score for each frame that conveys the positive/negative state label for the frame. The use of a hinge loss function is reported in \cite{tang2013deep} to produce excellent results on different classification problems. In addition to using hinge loss, we regularize the weights of the output layer using elastic net regularization, which is defined as adding a penalty term to the loss function $\ell(y, \hat{y})$, such that the function minimized becomes $\ell(y, \hat{y}) = \max (0, 1-y \cdot \hat{y}) + \lambda \big[ \alpha \lVert w \rVert_1 + (1-\alpha) \lVert w \rVert_2 \big]$, where $w$ is the set of weights, $\lVert\cdot\rVert_p$ is the $L_p$ norm, $\lambda$ is a penalty importance parameter, and $0 \leq \alpha \leq 1$.

To train this network, we first froze the weights of the non-modified layers so that they would not be changed. Training then proceeded using mini-batch stochastic gradient descent. Detailed information on training and results will be discussed in Section \ref{results_section}.

\section{Change-Point Detection}\label{change-point methods}

To recap our framework, we sample every $n$-th frame of a video and apply our classifier to each frame to distinguish between our states. This gives us a series of confidence scores. 
We then seek to find change-points in this series --- points at which the frames switch between our two states of interest. Due to the modular nature of our framework, we were able to explore several approaches to the problem of change-point detection.

\subsection{Change-Point Methods Overview}
Given a time series $X_i, i = 1...n$, we define a change-point $c$ as a place in the series where the underlying distribution of the $X_i$ changes. That is, in the case of one change-point:
\begin{center}
$X_i \sim F_1\ \forall\ i \leq c,\ X_i \sim F_2\ \forall\ i\ > c$
\end{center}
for some distributions $F_1 \neq F_2, c \in \{1...n\}$.  In the context of this problem, the distributions $F_1, F_2$ are over all frames representing our two states of interest. There may be zero, one, or multiple change-points in a given series of scores.

In this section, we discuss a variety of approaches to change-point detection. They are: 
\begin{itemize}
\item Mean squared-error minimization, for which we derive a distribution for its central statistic
\item Forecasting methods, which can be easily adapted to online change-point detection
\item Hidden Markov models and maximum likelihood estimation, which provide state labels for each frame
\end{itemize}

\subsection{Mean Squared Error minimization (MSE)}

We will first outline how this method (inspired by \cite{variation_com}) works for sequences with one change-point, then show how we extend it to address the possibility of multiple change-points.

In a single change-point sequence, we attempt to optimally describe the sequence by using two constant functions. That is, we find the optimal point to split the sequence such that the the two halves of the sequence cluster closely around their sample means. Formally:

\begin{equation*}
\begin{aligned}
MSE(c) &= \sum\limits_{i=1}^{c} \big( x_{i} - \bar{x}_{1} \big)^2 + \sum\limits_{i = c+1}^{n} \big(x_{i} - \bar{x}_{2} \big)^2,\\
\end{aligned}
\end{equation*}
\noindent
where

\begin{equation*}
\begin{aligned}
\bar{x}_{1} = \frac{1}{c} \sum\limits_{i=1}^{c} x_{i}, \ \bar{x}_{2} = \frac{1}{n-c} \sum\limits_{i=c+1}^{n} x_{i},
\end{aligned}
\end{equation*}

We are finding the total squared error from the two sample means. In the univariate case, we can reduce our expression to the following:

$$MSE(c) = \sum\limits_{i=1}^n x_i^2 + c\bar{x}_{1} ^ 2 + (n - c)\bar{x}_{2} ^ 2.$$

We now wish to create a hypothesis test to determine if a measurement of MSE at a given $c$ is significantly small enough to represent a change-point. Let $H_0$ be that $X_i$ does not have a change-point. We can reject this null hypothesis, and declare $H_1$ to be true (i.e., a change-point exists), if the $p$-value for $MSE(c)$ is below some significance threshold $\alpha$. The $p$-value calculations are given below.

Under $H_0$, we assume all $x_i$ are independently and identically distributed according to some distribution $X$. By the central limit theorem, we can take sample means $\bar{x}_{1}$ and $\bar{x}_{2}$ to be normally distributed for large enough sample size, i.e. $\bar{x}_{1} \sim \mathcal{N}(\mu_x, \frac{\sigma_x^2}{c}),\ \bar{x}_{2} \sim \mathcal{N}(\mu_x, \frac{\sigma_x^2}{n - c})$, where $\mu_x$ and $\sigma_x^2$ are the mean and variance of the $X$, respectively. Without loss of generality, let us assume $\mu_x = 0$. Then, the squared normal variable $\bar{x}_{1}^2$ is from the gamma distribution $\Gamma(\frac{1}{2},  \frac{2\sigma_x^2}{c})$, and also $\bar{x}_{2}^2$ is from $\Gamma(\frac{1}{2},  \frac{2\sigma_x^2}{n - c})$. So by the properties of the gamma distribution, we have $c\bar{x}_{1}^2 \sim \Gamma(\frac{1}{2}, \frac{2c\sigma_x^2}{c}) = \Gamma(\frac{1}{2}, 2\sigma_x^2)$ and $(n - c)\bar{x}_{2}^2 \sim \Gamma(\frac{1}{2}, \frac{2(n - c)\sigma_x^2}{n - c}) = \Gamma(\frac{1}{2}, 2\sigma_x^2)$. Therefore, $c\bar{x}_{1} ^ 2 + (n - c)\bar{x}_{2} ^ 2 \sim \Gamma(1, 2\sigma_x^2)$. Let us call this variable $G_c$. We then have
\begin{equation*}
\begin{aligned}
MSE(c) &= \sum\limits_{i=1}^n x_i^2 + G_c\\
MSE(c) - \sum\limits_{i=1}^n x_i^2 &= G_c \sim \Gamma(1, 2\sigma_x^2).\\
\end{aligned}
\end{equation*}

Since $\sum\limits_{i=1}^n x_i^2$ is a constant for each sequence, we can now calculate a $p$-value for $MSE(c)$, using the cumulative distribution function (CDF) for $\Gamma(1, 2\sigma_x^2)$. The CDF is as follows:

\begin{equation*}
\begin{aligned}
CDF(x) &= \frac{1}{\Gamma(k)} \gamma(k, \frac{x}{\theta}) 
&= \frac{1}{\Gamma(1)} \gamma(1, \frac{x}{2\sigma_x^2})
&= \frac{1}{1} \int\limits_0^\frac{x}{2\sigma_x^2} t^0 e^{-t} dt
&= -e^{-\frac{x}{2\sigma_x^2}} + 1\\
\end{aligned}
\end{equation*}
where $\Gamma(k)$ is the gamma function, and $\gamma(s, x)$ is the lower incomplete gamma function. Therefore, $p-value(x) = 1 - CDF(x) = e^{-\frac{x}{2\sigma_x^2}}$. Now, we can reject this null hypothesis if $p = e^{-\frac{x}{2\sigma_x^2}} < \alpha$, for significance level $\alpha$. When testing every point in a sequence, we can use a Bonferroni correction, with a new significance level of $\frac{\alpha}{n}$.

To find a single change-point, we find $MSE(c)$ for all $c$, and pick the one with the lowest p-value. If that p-value is below our threshold, that is the change-point; otherwise, we declare the sequence to be change-point free.

We can then recursively extend this method to find multiple change-points in a sequence, if they exist. We first find a single change-point --- as described above. If that change-point is deemed significant, we recursively test the intervals on each side of the change-point for another change-point. When an interval is deemed to not have a significant change-point, the algorithm stops.

\subsection{Forecasting Methods}
Forecasting methods allow us to fit a model to a set of data and then predict future observations using this model. To take advantage of the power of forecasting in the change-point detection setting, we develop what we will call the ``future window technique" (inspired by \cite{takeuchi2006unifying}) and combine it with univariate and multivariate modeling methods to detect change-points in the time series of frames. 

To employ the future window technique, we establish an initial model (or ``baseline model'') based on a set number of observations in the beginning of a time series --- assuming that a change will not occur within the first few observations of the time series.  To find potential changes, we use the baseline model to predict the next observation in the series and compare this prediction against a set number of future observations --- call this the ``future window." Comparing this prediction to multiple observations in the future allows us to see if the series deviates from the established model for a significant amount of time, which reduces instances of false positives created by outliers.  The number of observations in the window can be changed depending upon the desire of the user to either minimize false positives or false negatives.  If the differences between the prediction and each observed value in the window are all greater than some threshold --- whose determination differs from method to method --- then we call the value at the beginning of this window of observations a change-point.  If a change-point is established, we reestimate the baseline model using the observations in the future window \cite{takeuchi2006unifying}.  

The process outlined above repeats for every point in the time series, except for the last few where it would have been impossible to take a full future window into account. This methodology of estimating future observations based on a current model lines up with the framing of the change-point problem, which assumes that there is a shift in the model after a change-point, and the methodology enables the handling of cases where there are multiple change-points and cases where there are no change-points.  Furthermore, with minor modifications (which are not discussed in this paper), the future window technique could handle situations where-in-which the user does not have the entire data set all at once but, instead, receives pieces of data over-time.   

\subsubsection{Univariate Forecasting Methods}
For the following univariate methods, we assume that --- between change-points --- frames close to each other are temporally related and, thus, the SVM or CNN output for the frames is stationary 
\cite{usingeconometrics}.  Furthermore, we utilize the future window technique in conjunction with each of these univariate models; all of the values in the future window are used to re-estimate the model when a change-point is found.

We first utilize a one-lag autoregressive model, which accounts for the correlation between values in a time series by predicting the next value in the series on the basis of the previous observation.
Our threshold for the future window technique is the standard deviation of the entire time series \cite{psu_ar, forecastingbook}. 
Next, we combine the future window technique with a mean model, which computes the mean for a set number of observations and compares the values in the future window against this mean. We again use the standard deviation of the time series as the threshold for the future window technique \cite{dukemean}. 

Finally, we develop what we will call the ``sign-change filter.'' Between each potential change-point identified by a univariate algorithm, this filter computes the average of the CNN or SVM scores and then finds the sign of each average. If the sign of the average does not change at a potential change-point, we eliminate the change-point from the final output. This filter significantly increases the above methods' precision.

\subsubsection{Multivariate Forecasting Methods}

The BoVW histograms provide us with a succinct representation of a frame by counting the number of key-points (in a frame) associated with each visual word. We can apply methods directly to this time series, which is an unsupervised way of approaching the change-point detection problem.  We utilize histogram comparison methods in conjunction with the future window technique to accomplish this goal. We apply this methodology to the raw BoVW histograms and to condensed representations.  

We produce condensed representations by employing 
the agglomerate clustering algorithm to group the BoVW centroids that are within close proximity to each other \cite{jiang2009visual}. 
For our specific application, we apply the algorithm to the first one-hundred centroids, which represent features corresponding to the negative state, and then separately to the second set of one-hundred, which represent features corresponding to the positive state. 
After constructing the cluster tree, we choose the clusters of visual words such that we simplify the histograms without significantly reducing their informational content, and we achieve this by choosing an inconsistency coefficient cutoff.
For each histogram, we use the our chosen clusters to aggregate the key points into new ``bins'' \cite{hierarch_matlab}. 

To handle these multivariate representations, we utilize the chi-squared goodness-of-fit test and the match distance in conjunction with the future window technique. For each of these histogram comparison methods, we set the first observation in the series as the ``baseline model'' and, when we find a change-point, we set the new ``baseline model'' as the histogram in the beginning of the future window. 

The chi-squared goodness-of-fit test computes the squared differences between the bins of two histograms (call one the ``observed'' and one the ``expected'') and, for each bin, divides the squared difference by the number of elements in the ``expected'' histogram's bin, as demonstrated below 
$$ \chi^2 = \sum\limits_{i=1}^{k} \frac{(o_{i}-e_{i})^2}{e_{i}}, $$
where $o_{i}$ is the number of elements in the $i$-th bin of the ``observed'' histogram, $e_{i}$ is the number of elements in the $i$-th bin of the ``expected'' histogram, $k$ is the number of bins, and $k-1$ is the degrees of freedom. A $p$-value is computed for the resulting chi-squared value and compared to an alpha value, with the null hypothesis stating that the two histograms are similar and rejection of the null hypothesis indicating they are not. The expected histogram is the baseline histogram, and the observed histograms are the histograms in the future window. The threshold for the future window technique is the alpha level for the test so, if the $p$-values associated with the chi-squared values of the histograms in the future window are all less than alpha, we declare a change-point at the beginning of the window \cite{probbook,rubner2000earth}. 

The match distance finds the cumulative sum for each of two histograms, finds the absolute difference between the two sequences of partial sums, and then sums this resulting sequence.  This can be summarized by the equation:
$$ d_{M}(H,K) = \sum_{i = 1}^{n} | h_{i} - k_{i} |, $$
where $n$ denotes the number of bins and $h_{i}$ is the cumulative sum of the elements of $h$ up until and including bin $i$ \cite{rubner2000earth}. We find the match distance between the baseline histogram and each of the histograms in the future window. To set our threshold for the future window technique, for each feature/bin, we find the mean of the differences between successive observations in the series; we then sum these means and multiply this value by a constant. Both of these histogram methods were applied to the raw histograms and the condensed histograms. 


\subsection{Hidden Markov Model}

\begin{figure}[h]
    \centering
    \includegraphics[scale=0.3]{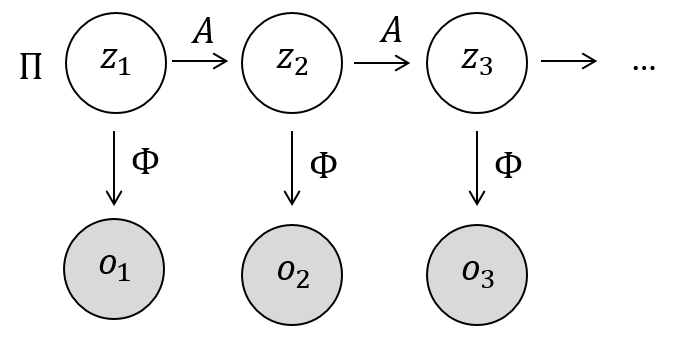}
\caption{Structure of HMM}
\label{hmm}
\end{figure}

In a hidden Markov model (HMM), the system being modeled is a sequence of discrete latent states, which, in our project, are the ground truths of whether the frames corresponds to the positive state. 
Such sequence is modeled using a Markov chain, in which the conditional probability of the future state only depends on the present state. Each type of state has an associated emission probability, according to which an output is assumed to be generated. While each latent state is not directly observable, the associated output is observable. Our goal was to construct the most probable sequence of latent states, given the sequence of associated classifier scores. 
Let $\{z_n\}^{N}_{n=1}$ be a sequence of latent states, where $z_n=[1\quad 0]^T$ if the frame corresponds to the negative state and $z_n=[0\quad 1]^T$ otherwise. Let $\{o_n\}^{N}_{n=1}$ be the associated sequence of classifier scores. The initial distribution $p(z_1)$ is given by $\pi=[\pi_1 \quad\pi_2]$, so that 
\begin{equation*}
p(z_1)=\begin{cases}
\pi_1\quad\textrm{if}\quad z_1=[1\quad0]^T\\
\pi_2\quad\textrm{otherwise}.
\end{cases}
\end{equation*}
The transition matrix of latent states is denoted as $A$, where $A_{ij}=p(z_{n,j}=1|z_{n-1,i}=1)$ and $i,j \in\{1,2\}$. We modeled the conditional distributions of observed variables using Gaussian distribution:
\begin{equation*}
p(x_n|z_n,\Phi)=\left(\frac{1}{\sqrt{2\pi\sigma_1^2}}\exp(-\frac{1}{2\sigma_1^2}(x_n-\mu_1)^2)\right)^{z_{n,1}}\left(\frac{1}{\sqrt{2\pi\sigma_2^2}}\exp(-\frac{1}{2\sigma_2^2}(x_n-\mu_2)^2)\right)^{z_{n,2}},
\end{equation*}
where $\Phi=\{\sigma_1,\sigma_2,\mu_1,\mu_2\}$ is the set of emission parameters. The structure of HMM is illustrated in Figure ~\ref{hmm}.

Estimates of parameters $\pi$, $A$, and $\Phi$ were computed by applying Expectation-Maximization (EM) algorithm and the Baum-Welch algorithm \cite{baum1972equality}. After this step, the most likely sequence of latent states were inferred using the Viterbi algorithm \cite{viterbi1967error}.

HMM is an unsupervised method, in which labels are not required for training. In principle, one can estimate parameters and then use these estimates to infer the most probable sequence of states using the same sequence of scores. In this project, however, we are more interested in evaluating how well a trained HMM can generalize to a new video. Our training and testing data sets were designed in the following way.

Videos with at least one exit or entrance into five folds are split into five folds. A HMM model is trained on four folds, and we apply a Savitzky-Golay filter \cite{savitzky1964smoothing} on the sequences of scores corresponding to videos in the remaining folds. The filtered sequences of scores are input into the trained HMM model. We declare a change-point occurs if two adjacent latent variables are inferred to have different states. This process is repeated five times, with each fold being the testing set exactly once.

In another experiment, the goal is to estimate the precision of HMM on all videos, including those without exit or entrance. To test HMM on videos without actual change-points, we apply a HMM trained on all videos that contain at least one actual change point. The results are presented in Table~\ref{tbl:multiple_change_point_univariate_all}.

\subsection{Maximum Likelihood Estimation}

Solving problems through maximum likelihood estimation is a common technique in the machine learning literature. The goal in maximum likelihood estimation is to find the values of some parameters which are most likely given the data available. Here, we develop a maximum likelihood formulation of the change-point detection problem, where the parameters we are trying to find are the true state labels, $L_i$. Let $L_i \in \{0, 1\}$ be the ground truth labels 
of a series and $x_i \in \{0, 1\}, i \in {1 ... n}$ be the 
labels from a classifier with accuracy $p$. Then, we can find the log-likelihood of a series of labels given the data as follows.

\begin{equation*}
\begin{aligned}
\log\mathcal{L}(L| X) &= \log P(X | L)\\
&= \log\prod\limits_{i=1}^{n} P(X_i | L_i)\\
&= \log\prod\limits_{i=1}^{n} p^{I[x_i = L_i]}(1 - p)^{I[x_i \neq L_i]}\\
&= \sum\limits_{i=1}^{n} {I[x_i = L_i]}\log(p) + {I[x_i \neq L_i]}\log(1 - p)\\
&= \log(p)\sum\limits_{i=1}^{n} {I[x_i = L_i]} + \log(1 - p)\sum\limits_{i=1}^{n}{I[x_i \neq L_i]}\\
\end{aligned}
\end{equation*}

Using integer programming (IP), we maximize this quantity. The essential addition in the IP formulation is a constraint $M$ on the number of change-points allowable; otherwise, the algorithm will not be robust to any sort of noise. The IP optimizes over values of L as follows:
\begin{equation*}
\begin{aligned}
\text{maximize} \quad &\log(p)\sum\limits_{i = 1}^n I[x_i = L_i] + \log(1 - p)\sum\limits_{i = 1}^n I[x_i \neq L_i]\\
\text{subject to} \ &\sum\limits_{i=1}^{n-1} |L_i - L_{i+1}|  <\ M\\
&L_i \in \{0, 1\}\\
\end{aligned}
\end{equation*}
The expression to maximize is the likelihood, the first constraint limits the number of change points, and the second constraint ensures that there are only two labels, one for each state. The results can be found in Table~\ref{tbl:multiple_change_point_univariate} and Table~\ref{tbl:multiple_change_point_univariate_all}.

\section{Experimental Results}\label{results_section}
We now present the results of applying our framework to a data set provided by the LAPD. We define change-points in this data set as the places where an officer exited or entered a vehicle. Our two states of interest are inside and outside a vehicle, and being outside of a car corresponds to the positive state in our framework. These change-points are important because police-public interactions often occur when officers are outside of their vehicles. 

\subsection{Data Set Description}
Our data is provided by LAPD, from their BWV pilot program in Los Angeles' Central Division in 2014-2015. The body-worn videos were recorded using cameras that have roughly a 130\degree \, field-of-view, a resolution of 640x480, and a fisheye lens.  All videos are from the officer's point-of-view, as body cameras are mounted on officers' chests.

There are 691 videos in our data, with an average length of 9 minutes. 420 of these videos contain at least one change-point of interest (either a vehicle entrance or exit), up to a maximum of 11. Of these videos, 270 of them begin from the driver's side, 176 are during the nighttime, and in 274 of them, the vehicle is moving at some point during the video. In addition, some videos contain occasional camera field-of-view occlusions from the officers' hands, arms, or clothing. The overall effect is that this data set is highly varied, and it presents many of the challenges that one might expect from real-world video data --- unclear images, rapid camera movement, extreme luminance and contrast differences, etc.

\subsection{Training and Testing Sets Description}\label{train_test_description}
Since SVM is more sensitive to redundancy in training set, we prepare different training and testing sets for SVM and CNN. 
SVM is sensitive to redundancy in training set because the learned decision boundary may be shifted in response to aggregated penalties imposed by repeated examples that lie on wrong side of margin. This poses a challenge to our project: as content of consecutive video frames are often highly correlated, these frames' representations are expected to be quite similar. We therefore manually select video frames that go into a data set for training and testing SVM to reduce the impact of redundancy in video data. 
Out of the 420 videos that contain at least one entrance or exit of a car, we take 200 of them and then randomly assign these 200 videos into 10 folds. For each of these selected videos, we select `in-car'' and ``out-car'' frames, and the resulting data set has 515 ``in-car'' frames and 529 ``out-car'' frames. In a trial, a SVM is trained using nine folds and tested on the remaining fold. This process is repeated ten times, and each fold is used as the testing fold exactly once. The testing accuracy on the ten trials are then averaged to give one estimate.

Training for the CNN proceeds by 10-fold cross-validation on the entire data set of 691 videos. First, videos are split into ten different folds. Then, from the videos, frames are extracted every one second, to form a data set of approximately 466,000 frames. No deletions or selections were made, and all frames are retained. During the training process, one fold is held out, and the CNN is trained on nine folds. Performance statistics are computed for each of the ten folds, and then averaged. Averaging is a valid way to combine performance statistics in our case, because the number of frames and in-car/out-of-car percentages are roughly equal across folds. 

\subsection{Classification Results}\label{classification results}
\begin{figure}
\label{sp}
\centering

\begin{subfigure}[t]{.45\textwidth}
\centering
\includegraphics[width=\linewidth]{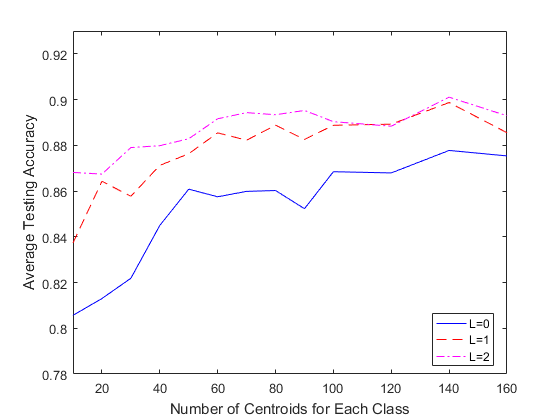}
        \caption{}\label{fig:fig_a}
\end{subfigure}
\begin{subfigure}[t]{.45\textwidth}
\centering
\includegraphics[width=\linewidth]{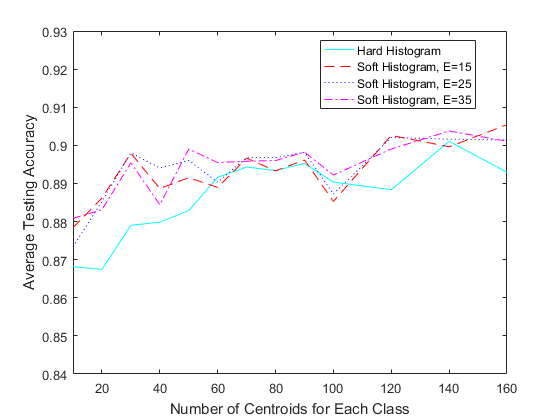}
\caption{}\label{fig:fig_b}
\end{subfigure}

\begin{minipage}[t]{1\textwidth}
\caption{Classification Accuracy of Support Vector Machine with Spatial Pyramid Match Kernel using hard histogram with different number of levels (\subref{fig:fig_a}), and using soft and hard histogram with different values of parameter $E$ (\subref{fig:fig_b}).}
\end{minipage}

\end{figure}

This section presents performance evaluations of our classifiers, SVM and CNN. Figure \ref{fig:fig_a} plots classification accuracy of SVM with spatial pyramid match kernel and hard histogram configuration versus number of clusters. The choice of $L$, which determines the total number of levels, has significant impact on classifier's performance. As $L$ increases from $0$ to $1$, which implies the spatial information of each keypoint is now taken into account, classifier's performance improves greatly. As $L$ increases from $1$ to $2$, each frame is partitioned into finer cells, and the extra spatial information also contributes to improvements in classification accuracy. The results also show that as the size of vocabulary increases, spatial information becomes less important. This observation is consistent with results in \cite{yang2007evaluating}.
Figure 
\ref{fig:fig_b} compares performance of SVM with hard and soft histogram configurations. To obtain these results, parameter $L$ is fixed at $2$ and we let parameter $E$ vary. As shown in the figure, soft VQ technique generally improve classification accuracy. For $E=35$, SVM with soft histogram outperform that with hard histogram at every size of visual vocabulary.

For CNN, the VGG-16 convolutional network architecture is modified for generalization, and to use the hinge loss function, as described in Section \ref{SS:CNN}. After this modification, all weights in all layers except for the last layer (the output layer) are frozen so that weight updates are not computed for them. For preprocessing, frames are then resized to 240x320, and the mean pixel value reported by the VGG-16 authors is subtracted from each color channel. The network is then trained via stochastic gradient descent with a mini-batch size equal to the size of the training set. Elastic net weight regularization, described in Section \ref{SS:CNN}, is used with $\alpha = 0.15$ and a penalty coefficient of 0.0003. The learning rate is initialized and scheduled according to an adaptive scheme, and decreases at every epoch. The network is trained for six epochs. Results are shown in Table \ref{tbl:nnres}. Implementation is carried out using the TensorFlow \cite{tensorflow2015-whitepaper}, Keras \cite{keras}, and scikit-learn software libraries \cite{scikit-learn}.

\begin{table}[h] 
\centering
\caption{Classification Results}
\begin{tabular}{cccc}
\hline
\textbf{Classifier}    & \textbf{Accuracy} & \textbf{Precision} & \textbf{Recall} \\ \hline
Best Convolutional Neural Network	& \textbf{94\%}	& \textbf{96\%}	& \textbf{95\%} \\
Best Support Vector Machine				& 90\%	& 92\%	& 89\% \\
\hline
\end{tabular}
\label{tbl:nnres}
\end{table}

The convolutional network results show the large improvement in performance statistics gained by using deep feature represenations of an frame (e.g.\ those computed by a ConvNet) as opposed to shallow feature representations (e.g.\ those computed by the BoVW process). We believe that more sophisticated training methods (such as jointly training a change-point detection method and CNN), or unfreezing the weights of the adapted VGG-16 network may be able to produce more accurate scene classifications.

\subsection{Change-point Detection Results}\label{change-point results}
In this section, we discuss the results of our change-point detection methods. As stated in the beginning of Section 4, we aim to identify points of vehicle entry and exit. For each video in our data set, we apply a classifier --- either CNN or BoVW-SVM --- to every $n$th frame (n = 30, 10 respectively). The classifiers output scores or class labels in \{0,1\}. We then run each sequence through each of five univariate change-point detection methods; for each sequence, we identify some number (possibly zero) of change-points.  We also run sequences of multivariate, unsupervised frame representations through multivariate change-point detection algorithms and, for each sequence, we identify some number (possibly zero) of change-points. 

To evaluate the performance of all of our change-point detection algorithms on our data set, we compare our algorithms' predicted change-points to the true change-points in the videos (where officers actually exit/enter their vehicles). We use a ten-second window of error, so a predicted change-point and a true change-point are considered equivalent if they are within ten seconds of each other. This accounts for the fact that it may take several seconds to exit or enter a vehicle. We then calculate precision and recall for each method to evaluate our performance --- where recall is the percentage of actual change-points which are within ten seconds of a predicted change-point, and precision is the percentage of predicted change-points which were within ten seconds of an actual change-point. These are aggregate measurements for all of the videos, meaning we count the total number of actual change-points and predicted change-points across all videos.

This section is organized as follows. First, we apply our methods on videos that contain at least one actual change-point using outputs from the CNN. 
Our algorithms are then tested on the full data set which contains videos without actual change points, and the results are jointly presented. We then discuss the results of running change-point detection algorithms on SVM outputs. Finally, we present the results of our change-point detection methods on multivariate, unsupervised representations.
As mentioned in Section \ref{train_test_description}, we undertake cross-validation to produce scores for all 691 videos using the CNN along with a traditional neural net classifier. Table \ref{tbl:multiple_change_point_univariate} shows the results of applying change-point detection algorithms on the CNN output for the 420 videos that contain at least one exit or entrance. While the five methods discussed in Section \ref{change-point methods} give comparable recall and precision, HMM produces the highest recall of 93\%, and MSE gives the highest precision of 75\%. 

We further test our change-point detection methods on CNN scores for the full data set containing 691 videos in total, 271 of which do not contain an entry into or an exit from a car. As shown in Table \ref{tbl:multiple_change_point_univariate_all}, recall calculations remain the same as those presented in Table \ref{tbl:multiple_change_point_univariate}, as these 271 videos do not contribute to the total number of actual change-points. Precision calculations, however, decrease because of false alarms.

For each of these methods, we can adjust some parameters. MSE uses a median filter window size of 30, a $p$-value cutoff of 0.1 with Bonferroni correction, and a maximum recursive depth of 3; it acts on the CNN binary labels. The autoregressive and mean model forecasting methods use the sample standard deviation of the series as the future window threshold, a future window of five, the first five observations to establish the baseline model, and the sign-change filter; they act on the CNN scores. MLE uses a parameter for classifier accuracy of 0.9, and a constraint on the number of allowable change-points of 10; it acts on the CNN binary labels. For HMM, the size and the polynomial order of the Savitzky-Golay filter are set to 15 and 1 respectively; this method acts on the CNN scores.



\begin{table}[H]
\centering
\caption[Univariate Multiple Change-Point Detection on Videos with Exit/Entrance]{Results of CNN Univariate Change-Point Detection on Videos with Exit or Entrance}
\label{tbl:multiple_change_point_univariate}
\begin{tabular}{ccc}
\hline
\textbf{Method}    & \textbf{Recall} & \textbf{Precision} \\ \hline
Hidden Markov Model       & \textbf{93\%}      & 72\%      \\ 
Mean-Squared Error Minimization      & 88\%      & \textbf{75\%}      \\ 
Forecasting Method -- Mean Model       & 88\%      & 70\%    \\
Maximum Likelihood Estimation Method       & 88\%      & 67\%      \\ 
Forecasting Method -- Autoregressive One Lag       & 85\%      & 70\%   
     \\ \hline
\end{tabular}
\end{table}

\begin{table}[H]
\centering
\caption[Univariate Multiple Change-Point Detection on all Videos]{Results of CNN Univariate Change-Point Detection on all Videos}
\label{tbl:multiple_change_point_univariate_all}
\begin{tabular}{ccc}
\hline
\textbf{Method}    & \textbf{Recall} & \textbf{Precision} \\ \hline
Hidden Markov Model       & \textbf{93\%}      & 65\%      \\ 
Mean-Squared Error Minimization       & 88\%      & \textbf{68\%}      \\ 
Forecasting Method -- Mean Model       & 88\%      & 61\%    \\
Maximum Likelihood Estimation Method      & 88\%      & 58\%      \\
Forecasting Method -- Autoregressive One Lag      & 85\%      & 60\%     \\
\hline
\end{tabular}
\end{table}

Table \ref{tbl:SVM_BoWV_output} presents the results of applying the change-point detection methods on the SVM scores. By comparing Table \ref{tbl:SVM_BoWV_output} with Table \ref{tbl:multiple_change_point_univariate}, we conclude that the precision measurements we calculate after running the change-point algorithms on SVM scores are significantly lower than the precision measurements we calculate after running the algorithms on CNN scores. Recall measurements, however, are generally comparable, except for MLE, whose recall decreases from 88\% to 66\%. From this piece of empirical evidence, we conclude that the performance of classifiers have a significant impact on the precision of change-point detection results.

Again, there are parameter values which we can adjust.  For the SVM output, the autoregressive forecasting method uses the sample standard deviation of the series as the future window threshold, a future window of five, and the first five observations to establish the baseline model; it acts on the SVM scores.  The mean model forecasting method uses the sample standard deviation of the series as the future window threshold, a future window of seven, the first ten observations to establish the baseline model, and the simple rounding filter (which rounds change-point values to the nearest thirty because of the way frames were sampled for the SVM); it acts on the SVM scores. MSE, MLE, and HMM use the same parameters as described above, but they act on SVM scores.

\begin{table}[H]
\centering
\caption{Univariate Change-point Detection Results on BoVW-SVM}
\label{tbl:SVM_BoWV_output}
\begin{tabular}{ccc}
\hline
\textbf{Method}    & \textbf{Recall} & \textbf{Precision} \\ \hline
Mean-Squared Error Minimization      & \textbf{91\%}      & \textbf{30\%} \\ 
Forecasting Method -- Mean Model        & 96\%      & 18\%    \\
Hidden Markov Model      & 90\%      & 17\% \\
Forecasting Method -- Autoregressive One Lag      & 90\%      & 17\%     \\
Maximum Likelihood Estimation Method     & 66\%      & 34\%      \\
\hline
\end{tabular}
\end{table}

Finally, Table \ref{tbl:multiple_change_point_multivariate_recall_precision} presents the results of change-point detection using multivariate data which primarily comes from the BoVW histograms.
These multivariate histograms represent the frames in their entirety, so they do not classify the frames into states --- unlike the scores and labels from the SVM and CNN.  Therefore, our methods may be detecting change-points in the video apart from exits from and entrances into vehicles.  Consequently, these methods may have fairly low precision values because they are detecting other changes besides car exits and entrances, and our precision measurements are just concerned with the car exit and entrance change-points. It is also worth noting that, for our condensed histogram representations, we yield similar results as the full histogram results recorded in Table 4.  Table 4 results are slightly better than results obtained using the condensed representations but, nevertheless, we realize that the histograms can be simplified without large losses in recall and precision.  

As with our univariate methods, our multivariate methods have some parameter values which we can adjust. MSE uses the same parameters as outlined in the univariate results section. The chi-squared test uses an alpha level of 0.001, a future window of seven, and the baseline for comparison as the first histogram.  The match distance uses a constant of 20 times the threshold discussed in Section 3.3, a future window of 10, the first histogram as the baseline, and the simple rounding filter (which rounds change-point estimations to the nearest thirty because of the way frames were sampled).

\begin{table}[H]
\caption[Results of Multivariate Multiple Change-Point Detection]{Results of Multivariate Change-Point Detection on Videos with Exit or Entrance}
\centering
\label{tbl:multiple_change_point_multivariate_recall_precision}
\begin{tabular}{ccc}
\hline
\textbf{Method}    & \textbf{Recall} & \textbf{Precision} \\ \hline
Chi-Squared Test & \textbf{100\%}      & \textbf{20\%}      \\ 
Match Distance       & 98\%      & 13\%    
     \\ 
Mean-Squared Error Minimization    & 86\%      & 17\%  \\ \hline
\end{tabular}

\end{table}



Above, we show both precision and recall for all our methods. In tuning our parameters, we prioritize recall over precision because, in a law enforcement application, we want to ensure to the best of our ability that we do not miss any important events. Our methods mostly achieve 85-90\% recall (MLE's recall is lower on SVM scores because it has fewer parameters to optimize, so we have less flexibility in choosing how we would like to manage the precision-recall trade-off.). The methods run on SVM scores yield a 15-35\% precision, and the methods run on CNN scores yield a 58-68\% precision. It is interesting to note the large discrepancy in change-point detection results for our different classification methods, despite a relatively small discrepancy (roughly 5\%) in their classification accuracy. It seems that a small improvement in classification performance can cause a large increase in precision of change-point detection.  Finally, for the CNN results in particular, we see that many of the methods yield quite similar recall and precision values.  This suggests there are multiple ways of approaching the change-point detection problem --- enabling a user to choose a method based on additional considerations such as algorithmic speed.

\section{Conclusion}

In this paper, we present a novel framework for change-point detection in video, using concepts from machine learning, image recognition, and change-point detection. We outline our methods for classification at the frame level, including CNNs and feature extraction techniques. We then describe methods from four approaches to change-point detection: mean square error minimization, forecasting, hidden Markov models, and maximum likelihood estimation. We present the performance of these methods on classifier output and on BoVW histogram representations. With our multivariate methods, we discuss the challenges of applying change-point detection methods to flexible, unsupervised, and multivariate representations. 

Testing specifically for identification of vehicle entrances and exits, our methods succeed with 90\% recall and nearly 70\% precision on a highly complex, realistic data set provided by the LAPD. However, we believe our framework is highly adaptable to different change-point classes, both within the domain of law enforcement BWV and outside of it. For instance, with an appropriately re-labeled data set, we believe our framework would succeed comparably at identification of video segments where an officer is speaking to a member of the public, handcuffing a suspect, or engaging in a foot chase.

With this in mind, the framework presented in this paper represents a promising step toward law enforcement's long-term goal of automatic video tagging of important segments. This would make the large-scale deployment of BWV (one camera to each officer in a police force) much more feasible.

\section{Acknowledgments}

First, we would like to thank our academic mentor, Dr. Giang Tran, for her guidance and continuous support. Her suggestions have helped us tremendously throughout this research. We would also like to thank Sgt. Javier Macias and Dr. Jeff Brantingham as our industry mentors; they have provided us with important context for this project. This work was completed as part of the 2016 Research in Industrial Projects for Students (RIPS) program at the Institute for Pure and Applied Mathematics (IPAM) and was supported by grants from the LAPD and NSF.

\bibliographystyle{siam}
\nocite{*}  
\bibliography{allBibliography.bib}

\end{document}